# Some Properties of Plausible Reasoning

Wray Buntine
RIACS*and AI Research Branch
NASA Ames Research Center, MS 244-17
Moffett Field, CA, 94025, USA
wray@ptolemy.arc.nasa.gov

## Abstract

This paper presents a plausible reasoning system to illustrate some broad issues in knowledge representation: dualities between different reasoning forms, the difficulty of unifying complementary reasoning styles, and the approximate nature of plausible reasoning. These issues have a common underlying theme: there should be an underlying belief calculus of which the many different reasoning forms are special cases, sometimes approximate. The system presented allows reasoning about defaults, likelihood, necessity and possibility in a manner similar to the earlier work of Adams. The system is based on the belief calculus of subjective Bayesian probability which itself is based on a few simple assumptions about how belief should be manipulated. Approximations, semantics, consistency and consequence results are presented for the system. While this puts these often discussed plausible reasoning forms on a probabilistic footing, useful application to practical problems remains an issue.

## 1 INTRODUCTION

There are many styles of knowledge representation and inference involving some form of uncertainty or inconsistency: reasoning about likelihoods, independence and related notions such as causality, confirmation, defaults and statistical frequencies, and tasks such as analogy, abduction and belief revision. In knowledge representation there is now a recognised need for the unification and development of these multiple, complementary forms of reasoning (Brachman, 1990). To do this unification an underlying belief calculus is needed as a common base for the different reasoning forms. First, this paper presents a system unifying defaults, likelihood, necessity and possibility. Because the system uses Bayesian probability as the underlying belief calculus, independence, abduction, belief revision and many other facets of plausible reasoning could be integrated as well, although it is not done so here. Second this paper argues that plausible reasoning can be interpreted as a form of approximate reasoning. This has important implications to the implementation of plausible reasoning systems. For instance, error can accumulate in a long chain of plausible reasoning, so potential error should be tracked. So this paper also discusses approximate methods for tracking error. Also, we cannot expect plausible reasoning to be correct every time. Many of the so-called paradoxes in plausible reasoning arise because it is assumed that plausible reasoning will always lead to a correct conclusion.

---

*Research Institute for Advanced Computer Science.

The system presented here, like most Bayesian methods, is based on a few basic assumptions together with a few approximations. The assumptions are about how belief can be modelled and updated and have been presented in (Horvitz et al., 1986). The same Bayesian principles have led to the development of algorithms for learning (Buntine, 1991a), uncertain inference, and many more applications outside of artificial intelligence. Bayesian methods are claimed to be normative, which means they set a standard for plausible reasoning and implies they will *not* suffer from the standard paradoxes that are discussed in the non-monotonic literature (Hanks and McDermott, 1987; Poole, 1989; Pearl, 1988; Etherington et al., 1990). (Treatment of several paradoxes are given here and in (Buntine, 1991b).) The system presented here only implements one facet of the Bayesian approach and therefore is incomplete and may require extending with one of many complementary modes of normative reasoning, such as the making of default assumptions about independence (Goldszmidt and Pearl, 1990b).

The next section informally introduces the notation for a qualitative and a quantitative logic that each demonstrate a different level of approximation for reasoning about defaults and likelihoods. The *default* component of the qualitative logic corresponds to the var-ious conditional logics developed for default reasoning (Delgrande, 1988; Pearl, 1988; Geffner, 1988) but most closely to Adams' improper conditional (Adams, 1966). The quantitative logic has its roots in remarks made by Adams. The qualitative system is presented here only for contrast with the quantitative system, because the quantitative system has many advantages with little extra overhead. The third section covers the theory of the two systems, semantics, consistency and consequence. Since the qualitative logic is an extension of Adams conditional logic (Adams, 1966; Adams, 1975), applied to default reasoning by Geffner and Pearl (Pearl, 1988; Geffner, 1988), this greatly extends and simplifies Adams' and Goldszmidt and Pearl's (Goldszmidt and Pearl, 1990a) consistency and consequence tests by incorporating necessity, possibility, and likelihood in a quantitative framework. The quantitative framework allows approximate default and likelihood reasoning and tracking of accumulated error at the same time. The fourth section illustrates the use of the logics on some standard problems from the literature. The fifth section uses the logics to illustrate some major properties of plausible reasoning.

It is beyond the scope of this paper to cover the basic notions of probability and decision theory underlying subsequent sections. Suitable introductions from an AI perspective can be found in (Langlotz and Shortliffe, 1989; Horvitz et al., 1988; Pearl, 1988).

## 2 NOTATION

$DP$ is a propositional[1] logic annotated with probability bounds, and has a probabilistic rather than a possible world semantics. This allows inequality reasoning as an approximation to normative reasoning about point probabilities. $QDP$ drops the numeric subscripts from $DP$ and is designed to be a qualitative counterpart of $DP$. It is intended to be an approximation to $DP$ for reasoning about "small" but not infinitesimal probabilities. The semantics of $QDP$ complements $DP$ and is based on order of magnitude reasoning, but also has an infinitesimal semantics similar to Adams' conditional logic.

$DP$ is built on the language $D_P$ that is constructed from the propositional language together with four modal operators: the unary connectives $\Box$ (necessity), $\diamond$ (possibility), and the binary connectives $\Rightarrow$ (default with error bound) and $\approx\!\succ$ (likelihood with lower bound). There is no nesting of these operators. The operators can be interpreted as follows. (Below $A$ and $B$ both represent arbitrary propositions.)

---
[1] Although propositional sentences are dealt with throughout, pseudo-first-order sentences will sometimes be used. They are effectively propositional if there are known to be a finite number of constants, no quantifiers are allowed, and a sentence with variables is intended to represent a sentence schema.

$\Box A$: $A$ is necessarily true in any situation.

$\diamond A$: Some situation can possibly arise in which $A$ is true.

$A \Rightarrow_\epsilon B$: Given that you know just $A$ about the current situation, you can infer $B$ by default (with error in belief at most $\epsilon$).

$A \approx\!\succ_e B$: Given that you know just $A$ about the current situation, $B$ is at least likely (with belief no less than $e$).

These four operators are joined using the standard boolean connectives ($\neg$ (negation), $\rightarrow$ (conditional), $\wedge$ (conjunction), etc.) to form the language $D_P$. This language also has a qualitative version, $QD_P$, which has the numeric subscripts dropped. The semantics for the language implements this by making $\epsilon$ and $e$ infinitesimal; not because we believe them to be infinitesimal but as a mathematical abstraction to obtain approximate behaviour of the operators for $\epsilon$ and $e$ small. $QD_P$ has successively weaker forms of the likelihood operator. $A \approx\!\succ B$ denotes "likely," whereas $A \approx\!\succ^2 B$ would denote "barely likely," etc. This is related to the iterated likelihood operator found in (Halpern and Rabin, 1987) and has a formal justification in Theorem 3 part 2.

$A \approx\!\succ^n B$: Given that you know just $A$ about the current situation, $B$ is at least likely to be ... to be likely (to order $n$).

The default and likelihood operators are "improper" according to Adams' terminology (Adams, 1966). This means $A \Rightarrow B$ and $A \approx\!\succ B$ will both hold true if $A$ is necessarily *false*. The "proper" versions must have $A$ being possible, so correspond to $\diamond A \wedge (A \Rightarrow B)$, and $\diamond A \wedge (A \approx\!\succ B)$ respectively.

The logics, being probabilistically based, are easily able to express sentences such as "an Australian is likely to drink Foster's": $Australian \approx\!\succ Drinks\text{-}Foster's$; whereas $Australian \approx\!\succ^2 Drinks\text{-}another\text{-}Foster's$ expresses the fact that, at least occasionly, an Australian will drink even more Foster's. Surprisingly enough, they also able to express sentences more in the spirit of autoepistemic (Moore, 1985) and default logics (Reiter, 1980). We can interpret the sentence "a professor has a Ph.D. unless known otherwise" two ways:

$$\diamond(Prof(x) \wedge Phd(x)) \longrightarrow (Prof(x) \Rightarrow Phd(x)),$$
$$\diamond(Prof(x) \wedge Phd(x)) \longrightarrow \Box(Prof(x) \rightarrow Phd(x)).$$

Read as "if it is possible that a particular professor has a PhD, then the professor *most likely* has a Ph.D.," and "if it is possible that a particular professor has a PhD, then the professor *definitely* has a Ph.D." respectively. The default logic representation, from $Prof(x) \wedge M\ Phd(x)$ infer $Phd(x)$, corresponds to the second reading. So the possibility operator, "$\diamond$", behaves rather like the $M$ operator of default logic.

# 3  THEORY

This section presents the semantics for the two logics and then discusses their intended use in plausible reasoning. Basic consistency and consequence theorems are given.

## 3.1  SEMANTICS

In $DP$, "$\models_{Pr} D$" denotes that $D \in D_P$ is true for the probability distribution $Pr$. $Pr$ plays a role not unlike an interpretation in standard propositional logic.

**Definition 1** *Given a probability distribution $Pr$ on propositions, "$\models_{Pr}$" is defined on sentences from $D_P$ as follows.*

1. *$\models_{Pr} \Box A$ if and only if $Pr(A) = 1$.*

2. *$\models_{Pr} A \Rightarrow_\epsilon B$ if and only if $Pr(B|A) \geq 1 - \epsilon$.*

3. *$\models_{Pr} \neg D$ if and only if not $\models_{Pr} D$.*

4. *$\models_{Pr} D \to E$ if and only if not $\models_{Pr} D$ or $\models_{Pr} E$.*

Possibility and likelihood are by definition dual operators for necessity and default respectively. "$\Diamond A$" is defined as "$\neg \Box \neg A$", so $\models_{Pr} \Diamond A$ if and only if $Pr(A) > 0$. "$A \not\Rightarrow_e B$" is defined as "$\neg(A \Rightarrow_e \neg B)$", so $A \not\Rightarrow_e B$ if and only if $Pr(B|A) > e$.

**Definition 2** *A sentence $D \in D_P$ is a theorem of the probabilistic logic $DP$ if $\models_{Pr} D$ for all possible probability distributions $Pr$.*

Consistency and consequence for sentences are defined in the usual manner based on the notion of a theorem.

To obtain qualitative rules about default and likelihood from the quantitative rules in $DP$, we can perform order of magnitude reasoning. We can consider a representative default error, $\epsilon$, where $\epsilon$ might be less than 0.01, or whatever the decision context requires. Likewise, we can consider a representative default likelihood, $e$, where $e$ might be greater than 0.05, say. In order to approximate the behaviour of our reasoning with these particular limits in mind, we can parameterise the system by $\epsilon$ and $e$ and consider only approximate calculations to $O(\epsilon)$ and $O(e)$. $QD_P$ is defined in a manner such that $\epsilon$ and $e$ are arbitrarily small, but $\epsilon$ is also arbitrarily smaller than $e$.

**Definition 3** *A sentence $D \in QD_P$ is a theorem of the qualitative probabilistic logic $QPD$ if there exists a theorem $D' \in D_P$ corresponding to $D$ (that is, identical except for any super or subscripts), in which all subscripts to "$\Rightarrow$" and "$\not\Rightarrow$" are parameterised by some variables $\epsilon$ and $e$ and each subscript to "$\Rightarrow$" is of order $\epsilon$ as $\epsilon$ approaches $0$ and $e$ remains finite, and each subscript in $D'$ corresponding to "$\not\Rightarrow^n$" in $D$ is of order $e^n$ as $e$ and $\frac{\epsilon}{e}$ approach $0$. This is denoted "$\models_{QDP} D$".*

Again, consistency and consequence are defined in the usual manner.

This definition can be reinterpreted to give an infinitesimal semantics close to that of Adams. Lemma 1 below (Buntine, 1991b) does this using a standard clausal form for defaults and another for likelihoods that collects all necessities and possibilities into the left-hand side of the clause.

**Lemma 1**

$\models_{QDP} \Box U \wedge_{i \in I_V} \Diamond V_i \wedge_{i \in I_A} A_i \Rightarrow B_i \longrightarrow \vee_{i \in I_C} G_i \Rightarrow H_i$ ,

*if and only if there exists a $\delta$ and $\eta$ such that for all $\epsilon < \eta$*

$\models_{DP} \Box U \wedge_{i \in I_V} \Diamond V_i \wedge_{i \in I_A} A_i \Rightarrow_\epsilon B_i \longrightarrow$
$$\vee_{i \in I_C} G_i \Rightarrow_{\delta\epsilon} H_i \ .$$

*Similarly,*

$\models_{QDP} \Box U \wedge_{i \in I_V} \Diamond V_i \wedge_{i \in I_A} A_i \not\Rightarrow^{n_i} B_i \longrightarrow \vee_{i \in I_C} G_i \not\Rightarrow^{m_i} H_i$ ,

*if and only if there exists a $\delta$ and $\eta$ such that for all $\epsilon < \eta$*

$\models_{DP} \Box U \wedge_{i \in I_V} \Diamond V_i \wedge_{i \in I_A} A_i \not\Rightarrow_{\epsilon^{n_i}} B_i \longrightarrow$
$$\vee_{i \in I_C} G_i \not\Rightarrow_{\delta\epsilon^{m_i}} H_i \ .$$

For the $D_P$ sentences in the lemma, $\delta$ is an *error propagation factor*, and $\delta\epsilon$ and $\delta\epsilon^{m_i}$ are the *error propagation functions* respectively. For the default clause, the larger the value of $\delta$, the faster error can propagate when the clause is applied in some chain of reasoning. Since a smaller likelihood represents more room for error, in the likelihood clause the smaller the value of $\delta$, the faster error will propagate when the clause is applied in some chain of reasoning.

For instance, the sentence

$$(A \not\Rightarrow_e C) \wedge (B \not\Rightarrow_d C) \to A \vee B \not\Rightarrow_f C \ ,$$

is a theorem of $DP$ with the error propagation function $f$ given by

$$f \leq \frac{ed}{e + d - ed} \leq \min(e, d) \ .$$

Therefore we can drop the subscripts to get a $QDP$ theorem as well.

## 3.2  THEOREMS

$DP$ and $QDP$ give a system for reasoning qualitatively and quantitatively about probability inequalities. However, normative reasoning according to Bayesian principles is based on point probabilities. Often in normative reasoning, we have a specific decision context in mind and we wish to determine if the probability of some proposition is less than or greater than some fixed probability (determined by the loss function). $DP$ and $QDP$ are then approximations for dealing with this special case. $QDP$ is merely an abstraction of $DP$ given here to show the connection of $DP$ with existing conditional and probabilistically motivated logics. Because of the inability of $QDP$ to keep track of error, it would be a potentially unsafe system to use in practice.

If the problem contains a good deal of uncertainty so the errors are large, or the loss function for the decisions to be made requires careful evaluation of comparative probabilities, it may be more appropriate to conduct a careful probabilistic analysis instead of using the approximate methods suggested here. If however, the errors are small, it is shown in this section we can do consistency and consequence tests in $DP$ using qualitative reasoning about defaults and likelihood, and follow this with some simple error propagation calculations to calculate upper bounds on propagated errors. These approximate probability calculations may then be a sufficient basis for making decisions. Details of this approach are described in this section. This makes $DP$ a safe alternative to $QDP$ when approximate reasoning seems appropriate.

Notice though that whether a sentence from $D_P$ is consistent or is a consequence of some other can be converted to a set of simplex problems in the variables, as done with Probabilistic Logic (Nilsson, 1986). We shall not pursue this approach, however, since we are concerned with approximate modelling of default and likelihood reasoning, for which "propagation errors" can be calculated rapidly using other more approximate means, as shown below.

Algorithms for consistency and consequence are given here for the numerically annotated logic $DP$. To obtain results for $QDP$, simply drop the subscripts, and in the case of likelihoods, be careful to check the orders of magnitude of the error propagation functions. Since each of the theorems below allows arbitrary possibilities to be included, the algorithms can be readily converted to the proper versions of the operators.

The algorithms rely on first computing the subset of the default (likelihood) operators that must have their antecedents necessarily false. For instance, in $(A \Rightarrow B) \land (A \Rightarrow \neg B) \land (C \Rightarrow D)$, $A$ must be necessarily false since both $B$ and $\neg B$ cannot be "typical" at the same time. So both the first two defaults must have their antecedents necessarily false. These computed subsets for defaults (likelihoods) are referred to as the maximum (minimum) inconsistent set. An algorithm for computing the maximum inconsistent set of a $DP$ sentence with defaults is given in Figure 1. The algorithm for computing the minimum inconsistent set for a $DP$ sentence containing no defaults is given in Figure 2. Logical tests for consistency and consequence are given in Theorem 2 for $D_P$ clauses containing no likelihood operator. The role of the maximum inconsistent set can best be seen by looking at

---

**Input:** A $D_P$ sentence $\Box U \land_{i \in I_V} \Diamond V_i \land_{i \in I_A} A_i \Rightarrow_{\epsilon_i} B_i$, where $\epsilon_i < \frac{1}{|I_A|}, \frac{1}{2}$ for $i \in I_A$.
**Output:** The maximum inconsistent set, $I_{max}$.
**Algorithm:** Let $I = I_A$. If there exists a $j \in I$ such that $U \land A_j \land_{i \in I} (A_i \rightarrow B_i)$ is satisfiable, then remove that $j$ from $I$. Repeat this until no $j$ found or $I = \emptyset$. $I_{max}$ is then given by $I$.

Figure 1: The defaults-inconsistency algorithm

---

**Input:** A $D_P$ sentence $\Box U \land_{i \in I_V} \Diamond V_i \land_{i \in I_A} A_i \approx_{\epsilon_i} B_i$, where $\epsilon_i < \frac{1}{|I_A|}$ for $i \in I_A$.
**Output:** The minimum inconsistent set, $I_{min}$.
**Algorithm:** Let $I = \emptyset$. If there exists a $j \in I_A - I$ such that $U \land_{i \in I} \neg A_i \land A_j \land B_j$ is unsatisfiable, then add that $j$ to $I$ and repeat until no $j$ found. $I_{min}$ is now given by $I$.

Figure 2: The likelihood-inconsistency algorithm

---

parts 1 and 4 of the theorem.

**Theorem 2** *Consider the $D_P$ sentence $D$ given by $\Box U \land_{i \in I_V} \Diamond V_i \land_{i \in I_A} A_i \Rightarrow_{\epsilon_i} B_i$, where $\epsilon_i < \frac{1}{|I_A|}, \frac{1}{2}$ for $i \in I_A$. Let $I_{max}$ denote the (unique) maximum inconsistent set for the sentence.*

1. *The sentence $D$ is inconsistent if and only if there exists some $j \in I_V$ such that $U \land V_j \land_{i \in I_{max}} \neg A_i$ is unsatisfiable.*

2. *The $D_P$ sentence $C \Rightarrow_\delta B$ is a consequence of $D$ for some $\delta < \frac{1}{2}$ if and only if $D \land (C \Rightarrow_\delta \neg B)$ is inconsistent. $\delta = \sum_{i \in I_A} \epsilon_i$ is a correct error propagation function.*

3. *The $D_P$ sentence $\Diamond C$ is a consequence of $D$ if and only if $D \land \Box \neg C$ is inconsistent.*

4. *The $D_P$ sentence $\Box C$ is a consequence of $D$ if and only if $D$ itself is inconsistent or $\models U \land_{i \in I_{max}} \neg A_i \rightarrow C$.*

Notice by part 1, if the $DP$ sentence contains proper default operators (so possibilities are included), then the sentence will necessarily be inconsistent if the maximum inconsistent set is non-empty. The corresponding property applies to likelihoods.

Tests for consistency and consequence using the likelihood operator are given in Theorem 3. Methods for computing tighter bounds for the error propagation function, linear in some cases, are given in (Buntine, 1991b).

**Theorem 3** *Consider the $D_P$ sentence $D$ given by*

$\Box U \wedge_{i \in I_V} \Diamond V_i \wedge_{i \in I_A} A_i \approx_{e_i} B_i$, where $e_i < \frac{1}{|I_A|}$ for $i \in I_A$. Let $I_{min}$ denote the (unique) minimum inconsistent set.

1. The sentence $D$ is inconsistent if and only if there exists some $j \in I_V$ such that $U \wedge_{i \in I_{min}} \neg A_i \wedge V_j$ is unsatisfiable.

2. The $D_P$ sentence $C \approx_f B$ is a consequence of $D$ for some $f < 1$ if and only if $D$ is inconsistent or the likelihood inconsistency algorithm in Figure 3 terminates yielding a consequence. If consequence holds, then a lower bound on $f$, the error propagation function, is given by $f \geq \left(\frac{e}{1+e}\right)^{|I_A|}$, where $e = \min_{i \in I_A} e_i$, although the error propagation function can be less, for instance, linear in the $e_i$ in some cases.

3. The $D_P$ sentence $\Diamond C$ is a consequence of $D$ if and only if $D \wedge \Box \neg C$ is inconsistent.

4. The $D_P$ sentence $\Box C$ is a consequence of $D$ if and only if $D$ is inconsistent or $\models U \wedge_{i \in I_{min}} \neg A_i \to C$.

---

**Input:** A consistent $D_P$ sentence $\Box U \wedge_{i \in I_V} \Diamond V_i \wedge_{i \in I_A} A_i \approx_{e_i} B_i$, where $e_i < \frac{1}{|I_A|}$ for $i \in I_A$, its minimum inconsistent set $I_{min}$, and a likelihood $C \approx_f D$.

**Output:** Whether the likelihood is a consequence of the sentence for some value of $f$.

**Algorithm:** If $U \wedge_{i \in I_{min}} \neg A_i \wedge C \wedge \neg B$ is unsatisfiable, return *is a consequence* for any $f$. Set $I = \emptyset$. If there exists some $j \in I_A - I_{min} - I$ such that

$$\models U \wedge_{i \in I_{min}} \neg A_i \wedge A_j \wedge B_j \wedge_{i \in I} (A_i \to B_i) \to C \wedge B$$

then add that $j$ to $I$. Otherwise return *not a consequence*. Repeat this process until $U \wedge_{i \in I_{min}} \neg A_i \wedge C \wedge \neg B \wedge_{i \in I} A_i$ is unsatisfiable or $I = I_A - I_{min}$. If termination occurred because $I = I_A - I_{min}$, return *not a consequence*, else return *is a consequence* for some $f$.

---

Figure 3: The likelihood-consequence algorithm

Also, $(A \approx B) \to \neg(A \Rightarrow \neg B)$ is a theorem of $QDP$. This property can be used, for instance, to convert a $QDP$ formula containing a mixture of defaults and likelihoods into a stronger formula containing just defaults, and so prove consistency of the weaker formula.

## 4  EXAMPLES

The logics are illustrated here on some standard paradoxes from the knowledge representation literature. Others handled are the "Yale shooting problem" and "Can Joe read and write?" (Buntine, 1991b).

### 4.1  THE LOTTERY PARADOX

Suppose a lottery has $1,000,000$ participants. The following two sentences are theorems of $DP$. The first follows from Theorem 2, and the second is its dual constructed by converting defaults to likelihoods and rearranging:

$$\bigwedge_{i=1}^{1,000,000} (true \Rightarrow_\epsilon \langle \text{person } i \text{ wont win lottery} \rangle) \longrightarrow$$
$$(true \Rightarrow_{1,000,000*\epsilon} \langle \text{no-one will win lottery} \rangle),$$

with its dual,

$$(true \approx_\epsilon \langle \text{someone will win lottery} \rangle) \longrightarrow$$
$$\bigvee_{i=1}^{1,000,000} (true \approx_{\frac{\epsilon}{1,000,000}} \langle \text{person } i \text{ will win lottery} \rangle).$$

Moreover, replacing $1,000,000$ by $999,999$ yields sentences that are not theorems of $QDP$. Ignoring the error bounds as done in $QDP$, the first sentence would seem to read "if, by default, any particular person will not win the lottery, then, by default, no-one will win the lottery at all". Likewise, the second $DP$ sentence would seem to read: "if it is likely that someone will win the lottery, then for some lottery entrant, it is likely they will win the lottery" (clearly not the case before the draw).

The two readings are versions of the lottery paradox that are the dual of each other. In the first $DP$ sentence the natural value for $\epsilon$ is $\frac{1}{1,000,000}$; this leaves the sentence impotent because the error bound in the conclusion becomes 1. In $DP$ there is no paradoxical reading. $QDP$ unfortunately drops the subscripts (both are of order $\epsilon$ as $\epsilon$ approaches 0) and loses the error information. $QDP$ suffers from the lottery paradox because it disregards the approximate nature of the default and likelihood operators. In the first sentence above, taking the conjunction of one million different approximate statements leads to an incorrect statement because the error in each accumulates.

Because of the cheap cost of maintaining approximate error calculations, as demonstrated in Theorem 3 for $DP$, there would seem little reason for using a purely qualitative system such as $QDP$.

### 4.2  THE "VANISHING" EMUS

The modelling of default reasoning based on infinitesimal probabilities has been criticised on the grounds that it makes "subclasses vanish" (Neufeld et al., 1990, p123). Etherington, Kraus and Perlis (Etherington et al., 1990) show a related problem applies to default logic and circumscription.

Consider the following rules:

$$Emu(x) \rightarrow Bird(x) ,$$
$$Emu(x) \Rightarrow \neg Flies(x) ,$$
$$Bird(x) \Rightarrow Flies(x) .$$

We can conclude (using Theorem 2) that "typically, birds aren't emus", $Bird(x) \Rightarrow \neg Emu(x)$, and "typically, things aren't emus", $true \Rightarrow \neg Emu(x)$.

If we take the infinitesimal semantics of the default operator literally then we could conclude that "no birds are emus", or "nothing is an emu". The real intent of the probabilistic semantics presented here, however, is about approximations so a more correct reading of the conclusion is that the emu is an uncommon or non-typical bird, which in reality is true of emus.

Circumscription, when presented with this same problem will deduce there are no emus to minimise the exceptions (Etherington et al., 1990). Etherington, Kraus and Perlis invent the notion of scope to overcome the same kind of difficulties in default logics and circumscription (Etherington et al., 1990):

> We contend that the intention of default reasoning is generally not to determine the properties of every individual in the domain, but rather those of some particular individuals of interest.

$QDP$ resolves the same paradoxes using a related principle that falls out naturally from the Bayesian framework and can be stated as follows:

> The intention of default reasoning is generally to determine reasonable properties of an individual in the domain. While these may be reasonable individually, they are not necessarily correct so one cannot reasonably say they apply uniformly.

## 5 CONCLUSION

The systems presented here do not do full normative Bayesian reasoning but instead are approximations valid in certain situations (as explained at the beginning of Section 3.2). Approximations have two effects: they can make a system incomplete or incorrect. $DP$ has retained correctness but become incomplete. In $QDP$ correctness is also lost by doing order of magnitude reasoning. One result of incompleteness is that on many general problems these systems will need complementary reasoning forms in order to produce a result. A result of incorrectness is that errors in reasoning can creep in, especially when they are hidden in qualitative reasoning which has a logical form making it appear deceptively accurate. As shown with the examples, both these results are a source of material for paradoxes if the underlying approximations are not understood.

This section discusses the issues raised by this: unifying complementary reasoning forms, the nature of approximate reasoning, and the dualities between default and likelihood reasoning. These insights, together with Theorems 2 and 3 form the major contributions of this paper. This gives us a much deeper insight into the problems of knowledge representation and inference involving some form of uncertainty.

### 5.1 DUALITIES

One of the first things taught to students of logic is the duality between disjunction and conjunction ($\neg(A \wedge B) \leftrightarrow (\neg A \vee \neg B)$ and $\neg(A \vee B) \leftrightarrow (\neg A \wedge \neg B)$). In modal logic, duality also holds between necessity and possibility ($\Box A \leftrightarrow \neg \Diamond \neg A$ and $\Diamond A \leftrightarrow \neg \Box \neg A$). In $DP$ the corresponding duality applies between default and likelihood. This means, for instance, that we can obtain dual forms for all $DP$ theorems and to a limited degree some $QDP$ theorems (the $QDP$ definitions are only approximately dual) by converting defaults to likelihoods and vice versa. Versions of some $QDP$ theorems and their (rearranged) duals are given in Table 1.

| 1 | $(C \Rightarrow A) \wedge (C \Rightarrow \neg A) \rightarrow \Box \neg C$ |
|---|---|
| 2 | $(C \Rightarrow A) \wedge (C \Rightarrow B) \rightarrow C \Rightarrow (A \wedge B)$ |
| 3 | $(C \Rightarrow A) \rightarrow (C \wedge A \Rightarrow B) \rightarrow (C \Rightarrow B)$ |
| 4 | $(A \vee B) \Rightarrow C \rightarrow (A \Rightarrow C) \vee (B \Rightarrow C)$ |
| 1 | $\Diamond C \rightarrow (C \succ\!\!\!\sim A) \vee (C \succ\!\!\!\sim \neg A)$ |
| 2 | $C \succ\!\!\!\sim (A \vee B) \rightarrow (C \succ\!\!\!\sim A) \vee (C \succ\!\!\!\sim B)$ |
| 3 | $(C \succ\!\!\!\sim B) \rightarrow (C \succ\!\!\!\sim A) \vee (C \wedge A \succ\!\!\!\sim B)$ |
| 4 | $(A \succ\!\!\!\sim C) \wedge (B \succ\!\!\!\sim C) \rightarrow (A \vee B) \succ\!\!\!\sim C$ |

Table 1: Some theorem schemata (1st table) and their duals (2nd)

These duality properties come about because of the basic properties of negation and by the dual definitions for the operators. A more remarkable but not so exact duality can be seen in the consistency and consequence theorems for default and likelihood. Compare the algorithms for the maximum and minimum inconsistent sets, and compare each of the results in Theorems 2 and 3. These theorems are not duals according to the definition of default and likelihood. For instance, the dual results to Theorem 2 would show a disjunction of likelihoods can be a consequence of a single likelihood rather than show a single likelihood can be a consequence of a conjunction of likelihoods, the situation of Theorem 3. The theorems are proven using quite different methods (for instance the results for likelihood are considerably harder to prove than those for default). Yet the theorems and algorithms have a remarkably similar form. Their major difference is that error combines slowly (linearly) for defaults but rapidly (multiplicatively) for likelihoods, though linearly in some special cases (Buntine, 1991b). Because likelihood errors combine rapidly, people often keep track of the degree of likelihood. For instance, likelihoods are used to rank order hypotheses in model-based diagnosis and abduction. Another result of this difference is that while considerable research has focussed on default reasoning, none to date has considered variable strength defaults as for instance allowed using error propagation functions and Theorem 2. In contrast, likelihood reasoning systems suggested in the literature introduced qualitative variable strength likelihoods from the beginning (Halpern and Rabin, 1987).

## 5.2 UNIFYING COMPLEMENTARY REASONING FORMS

The treatment of the two paradoxes "Can joe read and write?" and the Yale shooting problem are an example of how independence becomes an important complementary reasoning form for conditional logics. Both these problems yield no paradox in $QDP$, $NP$ (Delgrande, 1988) and related conditional logics because no default conclusions can be made at all. This holds because the antecedents of a conditional default or likelihood rule cannot be arbitrarily specialised with some additional knowledge. That is, the $QD_P$ sentence $(B \Rightarrow C) \to (A \wedge B \Rightarrow C)$ is not a theorem of $QDP$. For instance, the often useful transitive relation $(A \Rightarrow B) \wedge (B \Rightarrow C) \to A \Rightarrow C$ is not a theorem of $QDP$. However, the $QD_P$ sentence

$$((B \Rightarrow C) \to (A \wedge B \Rightarrow C)) \longrightarrow$$
$$((A \Rightarrow B) \wedge (B \Rightarrow C) \to A \Rightarrow C)$$

is a theorem of $QDP$. This means knowledge of the form $(B \Rightarrow C) \to (A \wedge B \Rightarrow C)$ will play a vital role in enabling default and likelihood conclusions like transitivity. If $A$ is independent of $C$ given $B$ then we have this knowledge.

Given that we need complementary reasoning forms, how do we unify them? It would be nice if we could somehow keep the different reasoning styles in separate modules, as suggested in hybrid reasoning systems (Frisch and Cohn, 1991). However, experience gained in the exercise here indicates this may not usually be possible. The unifying of necessity and possibility reasoning with default reasoning and likelihood reasoning, as presented in Theorems 2 and 3, required careful integration of the several approaches. Another unification that needs to be made is to integrate symbolic reasoning about independence (Lauritzen et al., 1990; Pearl, 1988) into the algorithms presented in Theorems 2 and 3.

## 5.3 APPROXIMATE REASONING

Qualitative reasoning about default and likelihood is interpreted here as an approximate form of reasoning that is bound to sometimes produce incorrect results. By investigating the quantitative counterpart to these reasoning forms, we are able to see more closely how this error propagates and accumulates and how we might track it, and we are able to better understand the assumptions under which the system operates. A qualitative system, for instance, has an implicit assumption that all errors $\epsilon$ are identical. With the quantitative system, however, we are able to allow the errors to vary—a more realistic situation.

Of course, all these rough approximations could be circumvented if we would adhere to more complete, fully normative Bayesian reasoning in the first place. This raises the important question: When do approximate systems such as $DP$ buy us improved performance in an application over more complete probabilistic approaches? Comparative studies here do not exist. Approximate systems such as $DP$ could be appropriate for generating a comprehensible explanation of probabilistic results obtained, for instance, by other numeric methods. Also, approximate systems due to their more simplistic framework, may be more appropriate for rapid turn-around in system development and user training. They may therefore serve as a useful complement to a more complete probabilistic approach rather than as a replacement. Only application experience will tell.

## References

Adams, E. (1966). Probability and the logic of conditionals. In Hintikka, J. and Suppes, P., editors, *Aspects of Inductive Logic*, pages 265–316. North-Holland, Amsterdam.

Adams, E. (1975). *The Logic of Conditionals*. Reidel, Boston.

Brachman, R. (1990). The future of knowledge representation. In *Eighth National Conference on Artificial Intelligence*, pages 1082–1092, Boston, Massachusetts.

Buntine, W. (1991a). Classifiers: A theoretical and empirical study. In *International Joint Conference on Artificial Intelligence*, Sydney. Morgan Kaufmann.

Buntine, W. (1991b). Modelling default and likelihood reasoning as probabilistic reasoning. *Annals of Mathematics and AI.* To appear.

Delgrande, J. (1988). An approach to default reasoning based on a first-order conditional logic: revised report. *Artificial Intelligence*, 36:63–90.

Etherington, D., Kraus, S., and Perlis, D. (1990). Nonmonotonicity and the scope of reasoning: Preliminary report. In *Eighth National Conference on*

*Artificial Intelligence*, pages 600–607, Boston, Massachusetts.

Frisch, A. and Cohn, A. (1991). Thoughts and afterthoughts on the 1988 workshop on principles of hybrid reasoning. *AI Magazine*, 11(5):77–83.

Geffner, H. (1988). On the logic of defaults. In *Seventh National Conference on Artificial Intelligence*, pages 449–454, Saint Paul, Minnesota.

Goldszmidt, M. and Pearl, J. (1990a). Deciding consistency of databases containing defeasible and strict information. In Henrion, M., Schachter, R., Kanal, L., and Lemmer, J., editors, *Uncertainty in Artificial Intelligence 5*. Elsevier Science Publishers, Amsterdam. An extended version appears as UCLA Cognitive Systems Laboratory, Technical Report CSD-890034 (R-122).

Goldszmidt, M. and Pearl, J. (1990b). A maximum entropy approach to nonmonotonic reasoning. In *Eighth National Conference on Artificial Intelligence*, pages 646–652, Boston, Massachusetts.

Halpern, J. and Rabin, M. (1987). A logic to reason about likelihood. *Artificial Intelligence*, 32:379–405.

Hanks, S. and McDermott, D. (1987). Nonmonotonic logic and temporal projection. *Artificial Intelligence*, 33:379–412.

Horvitz, E., Breeze, J., and Henrion, M. (1988). Decision theory in expert systems and artificial intelligence. *International Journal of Approximate Reasoning*, 2:247–302.

Horvitz, E., Heckerman, D., and Langlotz, C. (1986). A framework for comparing alternative formalisms for plausible reasoning. In *Fifth National Conference on Artificial Intelligence*, pages 210–214, Philadelphia.

Langlotz, C. and Shortliffe, E. (1989). Logical and decision theoretic methods for planning under uncertainty. *AI Magazine*, 10(1):39–48.

Lauritzen, S., Dawid, A., Larsen, B., and Leimer, H.-G. (1990). Independence properties of directed Markov fields. *Networks*, 20:491–505.

Moore, R. (1985). Semantical considerations on nonmonotonic logic. *Artificial Intelligence*, 25:75–94.

Neufeld, E., Poole, D., and Aleliunas, R. (1990). Probabilistic semantics and defaults. In Schachter, R., Levitt, T., Kanal, L., and Lemmer, J., editors, *Uncertainty in Artificial Intelligence 4*. North Holland.

Nilsson, N. (1986). Probabilistic logic. *Artificial Intelligence*, 28:71–87.

Pearl, J. (1988). *Probabilistic Reasoning in Intelligent Systems*. Morgan and Kauffman.

Poole, D. (1989). What the lottery paradox tells us about default reasoning. In *First International Conference on Principles of Knowledge Representation and Reasoning*, pages 333–340, Toronto.

Reiter, R. (1980). A logic for default reasoning. *Artificial Intelligence*, 13:81–132.